\documentclass[runningheads]{llncs}
\usepackage[T1]{fontenc}
\usepackage{graphicx}%
\usepackage{multirow}%
\usepackage{amsmath,amssymb,amsfonts}%

\usepackage{mathrsfs}%

\usepackage{xcolor}%
\usepackage{textcomp}%
\usepackage{manyfoot}%
\usepackage{booktabs}%
\usepackage{algorithm}%
\usepackage{algorithmicx}%
\usepackage{algpseudocode}%
\usepackage{listings}%
\usepackage{CJKutf8}
\usepackage{booktabs}  
\usepackage{siunitx} 
\usepackage{xurl}
\usepackage[hidelinks]{hyperref}

\begin{document}
\begin{CJK}{UTF8}{gbsn}

\title{Commercial Vehicle Braking Optimization: A Robust SIFT-Trajectory Approach}
%
%

\author{{Zhe Li}\inst{1,2} \textsuperscript{†},\and
{Kun Cheng}\inst{1,2}\textsuperscript{†}
\thanks{Corresponding author. E-mail: chengkun@ustc.edu} \and
{Hanyue Mo}\inst{1,2}\and
{Jintao Lu}\inst{1,2}\and
{Ziwen Kuang}\inst{1,2}\and
{Jianwen Ye}\inst{1,2}\and
{Lixu Xu}\inst{1,2}\and
{Xinya Meng}\inst{1,2}\and
{Jiahui Zhao}\inst{1,2}\and
{Shengda Ji}\inst{1,2}\and
{Shuyuan Liu}\inst{1,2}\and
{Mengyu Wang}\inst{1,2}}

\institute{College of Information Engineering, China Jiliang University, No. 258 Xueyuan Street, Hangzhou, 310018, Zhejiang, China \and Zhejiang-New Zealand Joint Laboratory on Vision-Based Intelligent Metrology,  China Jiliang University, Hangzhou 310018, China}

\authorrunning{F.Kun Cheng ,Zhe Li et al.}
%

\maketitle              
\begingroup
\renewcommand\thefootnote{}   
\footnotetext{† These authors contributed equally to this work.}
\endgroup

\begin{abstract}
A vision-based trajectory analysis solution is proposed to address the "zero-speed braking" issue caused by inaccurate Controller Area Network (CAN) signals in commercial vehicle Automatic Emergency Braking (AEB) systems during low-speed operation. The algorithm utilizes the NVIDIA Jetson AGX Xavier platform to process sequential video frames from a blind spot camera, employing self-adaptive Contrast Limited Adaptive Histogram Equalization (CLAHE)-enhanced Scale-Invariant Feature Transform (SIFT) feature extraction and K-Nearest Neighbors (KNN)-Random Sample Consensus (RANSAC) matching. This allows for precise classification of the vehicle's motion state (static, vibration,  moving). Key innovations include 1) multiframe trajectory displacement statistics (5-frame sliding window), 2) a dual-threshold state decision matrix, and 3) OBD-II driven dynamic Region of Interest (ROI) configuration. The system effectively suppresses environmental interference and false detection of dynamic objects, directly addressing the challenge of low-speed false activation in commercial vehicle safety systems. Evaluation in a real-world dataset (32,454 video segments from 1,852 vehicles) demonstrates an F1 score of 99.96\% for static detection, 97.78\% for moving state recognition, and a processing delay of 14.2 milliseconds (resolution 704×576). The deployment on-site shows an 89\% reduction in false braking events, a 100\% success rate in emergency braking, and a fault rate below 5\%.

\keywords{Zero-Speed Detection \and Automatic Emergency Braking \and Low-Speed False Activation \and Commercial Vehicle Safety \and Vision-Based}
\end{abstract}
\section{Introduction}\label{sec1}

Between 2016 and 2017, Great Britain recorded approximately 174,510 road casualties, including 27,010 fatalities or serious injuries (\cite{ref1}). The US Department of Transportation reports that more than 90\% of traffic accidents are the result of driver errors (\cite{ref2}). Connected and Automated Vehicles (CAVs) have transformative potential to improve driving safety, traffic flow efficiency, and energy consumption optimization (\cite{ref3}). As a critical active safety system, Automatic Emergency Braking (AEB) plays an important part in obstacle avoidance maneuvers (\cite{ref4}). But its practical application in commercial vehicles faces a persistent technical challenge: misactivation at low speeds (particularly in the range of 0-5 km/h), significantly compromising the reliability of the system and user acceptance (\cite{ref5}).

AEB operational logic is based on multiple decision criteria, with vehicle speed being the paramount. However, current implementations depend on Controller Area Network (CAN) signals that exhibit inherent precision limitations. This serial bus communication protocol, originally developed by Robert Bosch GmbH (\cite{ref6}), facilitates data exchange between ECUs but struggles to accurately represent velocities in the critical range of 0-5 km/h. The resulting "zero-speed braking" phenomenon occurs when erroneous CAN readings trigger unnecessary full brake applications despite actual vehicle motion.

Existing solutions for zero-speed detection primarily fall into two categories: velocity-based methods and Inertial Measurement Unit (IMU)-sensor fusion approaches (\cite{ref7,ref8,ref9}). This study introduces a cost-effective vision-based zero-speed detection algorithm as either a complementary or alternative solution. Leveraging blind-spot camera feeds, the proposed methodology employs adaptive feature analysis of sequential video frames to determine actual vehicle motion states. By integrating with the existing AEB decision pipeline through an auxiliary verification module, the system selectively suppresses unnecessary braking interventions only when true static conditions (including vibratory states) are confirmed, while maintaining robustness against environmental confounders such as illumination variations, precipitation, and dynamic object intrusions (e.g., pedestrians, vehicles).

Key innovations of this research include:
\begin{enumerate}
\item \textbf{Multi-frame trajectory displacement statistics}: Employing a 5-frame sliding window for temporal motion consistency analysis
\item \textbf{Adaptive CLAHE-SIFT feature enhancement}: Combining contrast normalization with scale-invariant feature extraction
\item \textbf{Dual-threshold state classification matrix}: Enabling precise differentiation between static/vibratory/moving states
\item \textbf{OBD-II-integrated dynamic ROI configuration}: Optimizing computational efficiency through region-of-interest adaptation

\end{enumerate}

\section{Related Work}
\subsection{Limitations of traditional AEB speed detection}
The IMU sensor solutions (\cite{ref8,ref9}) improved accuracy through acceleration compensation but suffered from vibration misjudgment, where engine idle vibrations were often mistakenly identified as motion. This study, for the first time, distinguishes between vibration and actual motion through visual trajectory displacement statistics, achieving an F1 score of 96.40\%(Figure III) for vibration state detection.

\subsection{Development of Visual Motion Detection}

The Scale-Invariant Feature Transform (SIFT) algorithm, proposed by David Lowe in 1999 and improved in 2004, extracts local feature descriptors that are invariant to scale, rotation, and illumination by constructing a difference-of-Gaussian pyramid and employing an extremum detection mechanism. It has become one of the most influential classic algorithms in the field of computer vision (\cite{ref10,ref11,ref12,ref13}). In the field of vehicle dynamics analysis, SIFT features are widely applied in core tasks such as motion estimation, target tracking, and multi-modal data fusion, owing to their unique robustness. Feature-based methods dominate vehicle motion analysis. Research (\cite{ref10}) indicates that Lowe's SIFT provides scale invariance but suffers from slow retrieval speeds and poor performance. This study utilizes  Contrast Limited Adaptive Histogram Equalization (CLAHE) to enhance the feature quality of low-contrast scenes and integrates CLAHE-SIFT, which adaptively enhances contrast in low-light conditions, thereby improving the success rate of feature matching.

\subsection{Challenges in Commercial Vehicle Scenarios}
Commercial vehicles face more severe environmental conditions compared to passenger vehicles. Research (\cite{ref14}) indicates that harsh weather conditions, such as rain and snow, can affect the accuracy of vision-based detection methods. Subsequent studies (\cite{ref15,ref16}) have improved robustness by integrating radar and cameras, but this approach increases costs. Our solution employs a dynamic ROI (Region of Interest) filtering mechanism to exclude non-road areas (e.g., truck side mirrors), thereby reducing unnecessary computations.

\subsection{Innovative positioning}
TABLE \ref{table1} compares the intergenerational differences between this scheme and existing technologies.

\begin{table}[htbp]
\centering
\caption{Technical Solution Comparison Table}
\begin{tabular}{|p{4cm}|p{2cm}|p{4cm}|p{3cm}|}
\hline
\textbf{Technical Dimension} & \textbf{Existing Solution} & \textbf{Our Innovation} \\ \hline
Feature Extraction & Baseline SIFT\ & CLAHE-SIFT + Dynamic ROI \\ \hline
State Classification & Binary (Moving/Static) & Ternary (Static/Vibration/\newline Moving) \\ \hline
Hardware &IMU+GPU Fusion & Monocular Camera + Edge Computing \\ \hline
\end{tabular}
\label{table1}
\end{table}

\section{Methodology}
\subsection{Overall Framework}
\begin{figure}[!h]
\centering
\graphicspath{ {./} }
\includegraphics[width=2.5in]{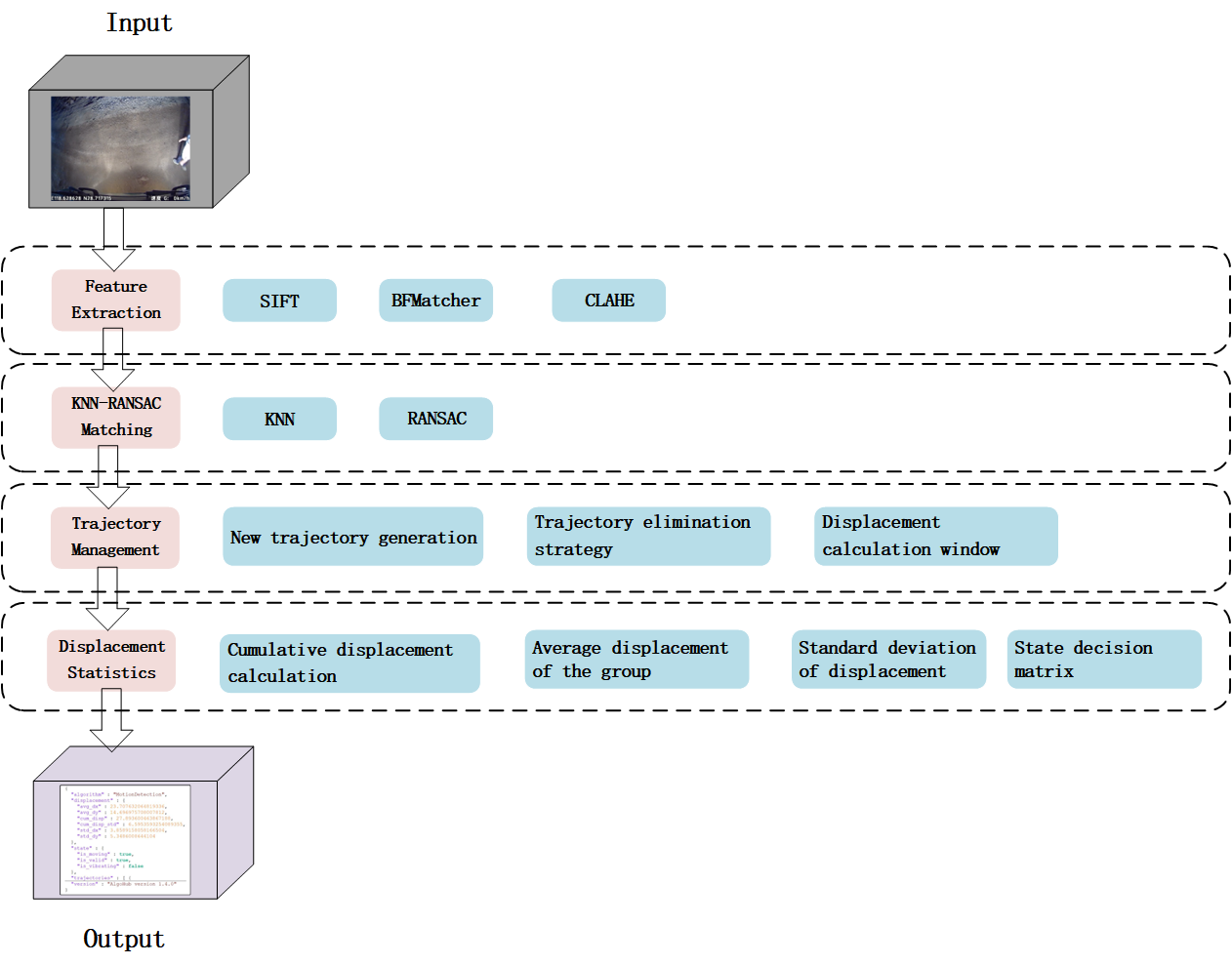}
\caption{Image Input to Output Processing Flow}
\label{fig1}
\end{figure}

\begin{figure}[h]
\centering
\graphicspath{ {./} }
\includegraphics[width=2.5in]{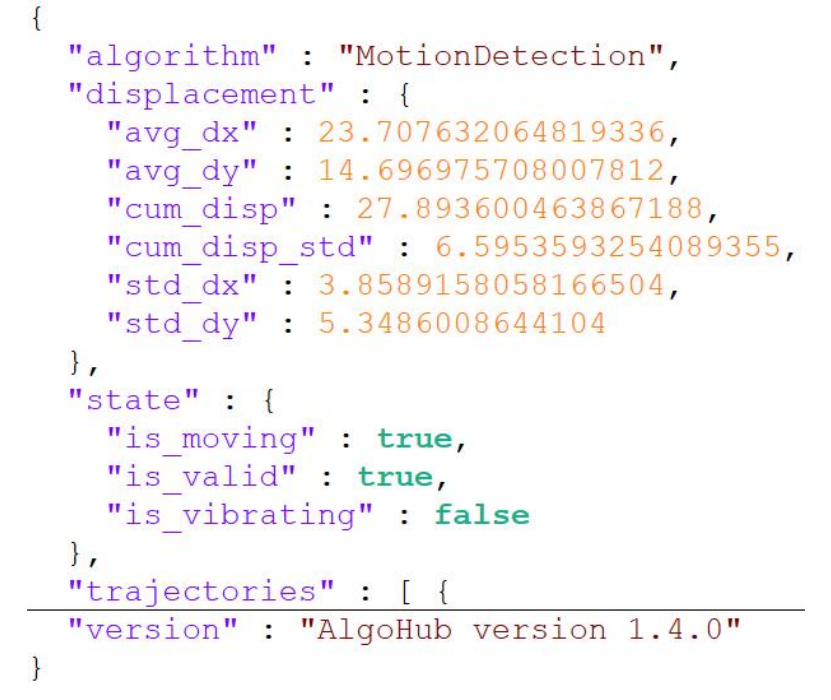}
\caption{Output JSON}

\label{fig2}
\end{figure}

The core algorithm framework of the system is shown in Figure \ref{fig1}. The general workflow of the algorithm is as follows:The system receives real-time video streams from the blind spot camera as input. Scale-Invariant Feature Transform (SIFT), Binary Feature Matching (BFMatcher), and Contrast Limited Adaptive Histogram Equalization (CLAHE) are employed to enhance image features. A Region of Interest (ROI) is defined to filter out irrelevant areas and reduce environmental interference. The feature points extracted by SIFT are then matched using the K-Nearest Neighbors (KNN) algorithm (\cite{ref18}), followed by verification through the Random Sample Consensus (RANSAC) algorithm to improve the accuracy and robustness of the matching process. Finally, through trajectory management and displacement statistics, the state of motion (static / vibrating / moving) is intelligently determined, and the result is packaged in JSON format, as shown in Figure \ref{fig2}, for the final output.

\subsection{Feature Extraction}
\begin{figure}[h]
\centering
\graphicspath{ {./} }
\includegraphics[width=2.5in]{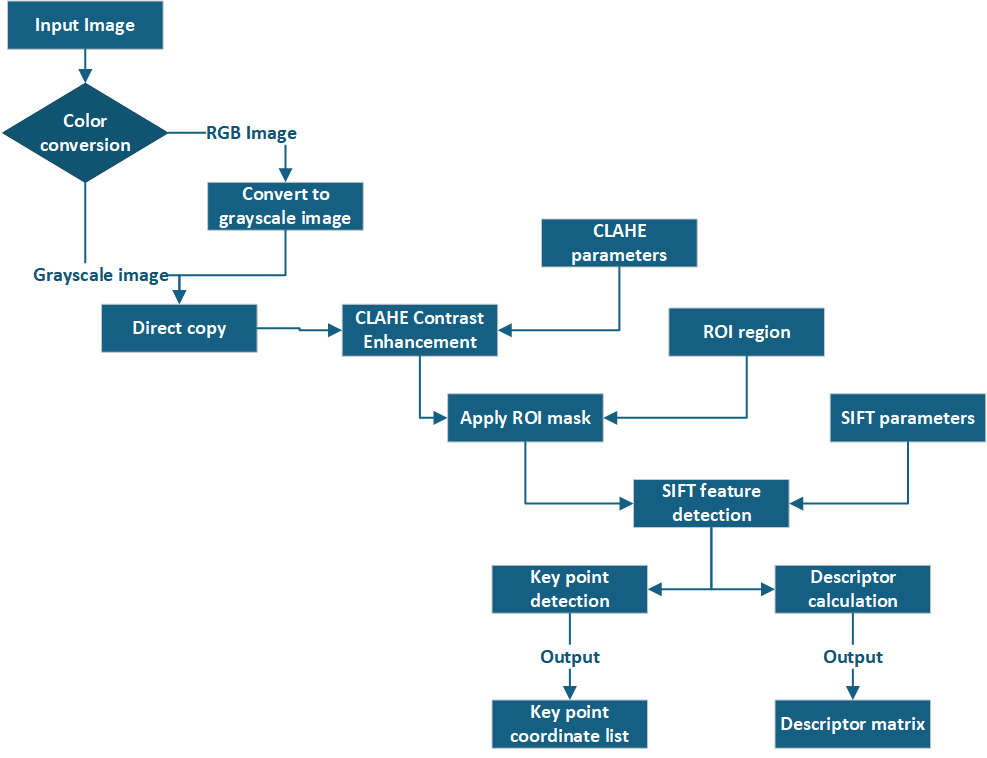}
\caption{Feature extraction module flowchart}
\label{fig3}
\end{figure}

Feature extraction (Figure \ref{fig3}), as one of the most important steps in image classification, can capture a certain visual attribute of the image (\cite{ref19}). Convert RGB images to grayscale when importing images to reduce computational complexity. Before feature detection, CLAHE (using an 8 × 8 grid histogram) is employed to enhance low-contrast regions. Dynamically set the focus area (top 5\% to bottom 90\%, left 15\% to right 86\%) through the ROI configuration file and adjust the actual size according to the installation requirements. In the Scale Invariant Feature Transform (SIFT) algorithms (\cite{ref11,ref20}), we set an upper limit of 1000 feature points to control computational complexity, a contrast threshold of 0.04 to filter out low-contrast noise points, and an edge response threshold of 15.0 to exclude edge instability points; when there is blur, high noise, or uneven lighting in the image, it is necessary to dynamically adjust the contrast threshold and edge threshold to enhance robustness.

\subsection{Trajectory Management}
Trajectory management involves four stages: creation, update, elimination, and reset. The system uses a 5-frame sliding window to maintain trajectories and determine states. The main technical implementation is as follows:
\begin{enumerate}
\item \textbf{Creation Mechanism:} When new feature points are successfully matched through KNN and RANSAC, the system assigns a unique ID to each point and initializes a trajectory queue, recording the initial position and the matching points of the current frame.

\item \textbf{Update Rule:} A sub-pixel level matching strategy is applied. Suppose the deviation between the current frame’s matching points and the endpoint of the previous frame’s trajectory is less than 1 pixel. In that case, the current point is appended, and the oldest position at the queue’s head is cleared.

\item \textbf{Trajectory Elimination Mechanism:} After processing each frame, if a trajectory has not matched in the two consecutive frames, it is immediately discarded. This mechanism effectively removes invalid trajectories caused by brief occlusions or mismatches.

\item \textbf{Sliding Window:} The system uses a 5-frame sliding window (kWindowSize=5) for trajectory updates and state determination, ensuring real-time and accurate trajectory management.
\end{enumerate}

\subsection{Displacement statistics stage}
\subsubsection{Cumulative displacement calculation}For each valid trajectory, this study calculated the displacement vector between its starting point (earliest position) and endpoint (latest position). The specific calculation method is as follows.
\begin{equation}
s= \sqrt{(x_i^{(end)} - x_{i}^{(start)})^2 + (y_i^{(end)} - y_{i}^{(start)})^2}
\end{equation}
Afterwards, collect displacement data from all valid trajectories and calculate two core statistics.
\subsubsection{Average displacement of the group}
The group average displacement is based on the matching results of feature points between consecutive frames and evaluates the motion state of vehicles by calculating the displacement of feature points on the image plane. The specific calculation formula is as follows:
\begin{equation}
\Delta dt = \frac{1}{n}\sum_{i=1}^{n} \sqrt{(x_i^{(end)} - x_{i}^{(start)})^2 + (y_i^{(end)} - y_{i}^{(start)})^2}
\end{equation}

Among them, n represents the total number of valid trajectories.
\(X_j^{end}\) and \(Y_j^{end}\) respectively represent the X and Y coordinates of the endpoint of the jth trajectory, and \(X_j^{start}\) and \(Y_j^{start}\) respectively represent the X and Y coordinates of the starting point of the jth trajectory. This study set a threshold kCumulativeThresh\_=2.05 pixels to compare with the actual calculated population average displacement to determine whether there is significant systematic motion.

\subsubsection{Standard deviation of displacement} To measure the consistency of group movement, this study calculated the standard deviation of X-axis displacement (σx) and Y-axis displacement (σy) and compared them with the set threshold kSdThresh\_=0.23. Firstly, calculate the average value of X-axis displacement (X) and Y-axis displacement (Y):

\begin{equation}
\bar{X} = \frac{1}{m} \sum_{j=1}^{m} X_j = \frac{1}{m} \sum_{j=1}^{m} \left( X_j^{end} - X_j^{start} \right)
\end{equation}

\begin{equation}
\bar{Y} = \frac{1}{m} \sum_{j=1}^{m} Y_j = \frac{1}{m} \sum_{j=1}^{m} \left( Y_j^{end} - Y_j^{start} \right)
\end{equation}

Then, calculate the standard deviation of X-axis displacement (σx) and Y-axis displacement (σy) according to the following formula:
\begin{equation}
\sigma_x = \sqrt{\frac{1}{m} \sum_{j=1}^{m} (X_j - \bar{X})^2}
\end{equation}
\begin{equation}
\sigma_y = \sqrt{\frac{1}{m} \sum_{j=1}^{m} (Y_j - \bar{Y})^2}
\end{equation}
Among them, \( x_{(j)} \) and \( y_{(j)} \) represent the displacement of the X axis and the displacement of the Y axis of the trajectory, respectively.

\subsubsection{State decision matrix}Based on the results of the trajectory displacement calculation, design a state decision matrix to determine the actual state of the vehicle's motion. The state decision matrix is based on the average displacement (Δd) and standard deviation (σ) of feature point displacements within the trajectory window. The specific classification criteria are as follows:

\begin{equation}
\text{Status} = 
\begin{cases} 
\text{Moving},  & \Delta dt > 2.05\,\text{px} \\
\text{Vibration},  & \Delta dt \leq 2.05\,\text{px} \ \wedge\ \sigma > 0.23\,\text{px} \\
\text{Static},  &  \Delta dt \leq 2.05\,\text{px} \ \wedge\ \sigma \leq 0.23\,\text{px}
\end{cases}
\end{equation} 
    
Using the decision matrix stated above and taking into account the different characteristics of vehicle motion, an accurate classification of vehicle motion states has been achieved. In practical applications, the matrix can be fine-tuned according to specific driving environments and vehicle characteristics.

\section{Experimental Verification}
\subsection{Test Configuration}
To evaluate the visual trajectory analysis scheme for commercial vehicle AEB systems, a dedicated dataset was constructed.

\textbf{Dataset}:  The surge in learning-based methods has significantly increased the demand for diverse training data (\cite{ref21}). Research (\cite{ref22}) mentions that early studies explored pure camera-based methods under adverse conditions (\cite{ref14,ref23}), but these methods used very small datasets, which are not suitable for commercial vehicle scenarios. Later, a series of multi-modal datasets (\cite{ref24,ref25,ref26,ref27}) focused on noise issues. For example, the GGROUNDED dataset (\cite{ref24}) emphasizes ground-penetrating radar localization under various weather conditions. The ApolloScape open dataset (\cite{ref26}) combines LiDAR, camera, and GPS data, including scenarios with rainy conditions and bright scenes. The Ithaca365 dataset (\cite{ref27}) is designed for the robustness of autonomous driving research, providing scenes under challenging weather conditions such as rain and snow. However, these datasets are not suitable for this study. Therefore, a dedicated dataset was created, composed of 32,454 video clips collected from 1,852 trucks, with each clip lasting 10 seconds. The dataset covers three motion states: static, vibration, and moving (slow and fast), including low-texture and high-texture road surface scenes. It covers various weather conditions, including sunny, overcast, and light rain, as well as changes in sunlight and shadows. This large-scale dataset covers most of the real-world scenarios encountered by commercial vehicles in operation. However, it does have some limitations, including insufficient extreme weather conditions, low video bitrate, and OSD interference.

\subsection{Implementation Details}
The motion detection system proposed in this study is based on a purely vision-based approach. The hardware platform utilizes the NVIDIA Jetson AGX Xavier (\cite{ref17}) as the embedded computing unit. Table \ref{table2} provides the detailed hardware specifications of the system.

\begin{table}[htbp]
\centering
 \caption{Detailed hardware parameters}
\begin{tabular}{|p{2cm}|p{6cm}|}
\hline
Device Type & Model/Parameters \\
\hline
Main Control Unit & Processor Intel(R) Core(TM) i9-14900KF 3.20 GHz \\
\hline
Main Control Unit & Onboard RAM 128 GB (128 GB available) \\
\hline
Main Control Unit & Storage 1.86 TB SSD GeIL P4A 2TB \\
\hline
Main Control Unit & Graphics Card NVIDIA GeForce RTX 4090 (24 GB) \\
\hline
Main Control Unit & System Type 64-bit Operating System, x64-based Processor \\
\hline
Vehicle Camera & APlus Blind Spot Camera, Uploaded Video is Compressed and Scaled \\
\hline
\end{tabular}
\label{table2}
\end{table}





\subsection{Results}
\textbf{Three types of classification results: }As shown in Table \ref{table3}, the Precision for the "moving" state reaches 99.985\%, the F1-score is 98.205\%, and the Recall is 96.486\%. However, the "static" state has a Recall of 73.6\%, with a Precision of only 65.151\%, a Recall of 73.603\%, and an F1-score of 69.12\%. Similarly, the "vibration" state has a Recall of 69.708\% and an F1-score of 69.497\%. The relatively lower scores for the static and vibration states are mainly due to feature overlap between these two states in low-contrast scenarios.

\begin{figure}[!h]
\centering
\graphicspath{ {./} }
\includegraphics[width=2.5in]{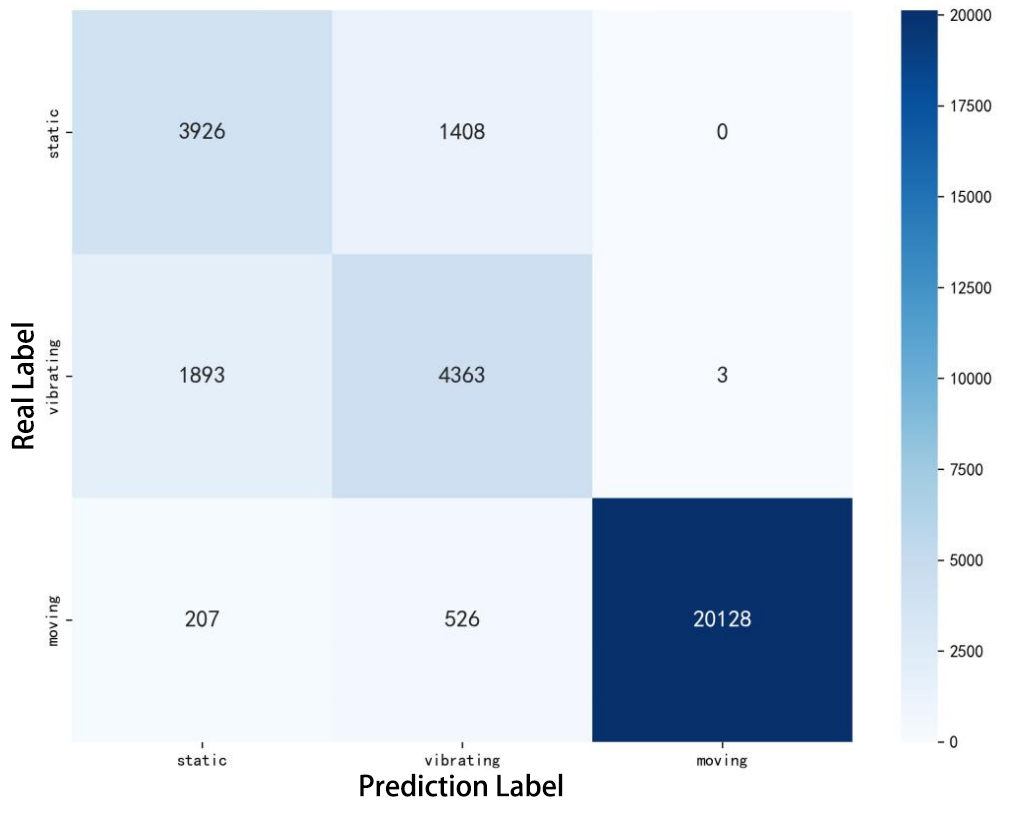}
\caption{Confusion matrix for classification of three types of motion states}
\label{fig4}
\end{figure}

The confusion matrix provides further insight into the specific classification performance of each category, as shown in Figure \ref{fig4}. A total of 3,926 static samples were correctly classified as static, but 1,408 were misclassified as vibration. This indicates that static and vibration states have similar features, making them prone to confusion.

\begin{table}[!h]
    \centering
    \caption{Performance analysis of three types of motion state classification details}
    \begin{tabular}{
        l               
        S[table-format=1.4]  
        S[table-format=1.4]  
        S[table-format=1.4]  
    }
    \toprule
    \textbf{State} & \textbf{Precision} & \textbf{Recall(Accuracy)} & \textbf{F1-score} \\
    \midrule
    Static & 0.65151 & 0.73603 & 0.69120 \\
    Vibration & 0.69287 & 0.69708 & 0.69497 \\
    Moving & 0.99985 & 0.96486 & 0.98205 \\
    \bottomrule
    \end{tabular}
    \label{table3}
\end{table}

\textbf{Two types of classification results:}
After merging the static and vibrating states into an unmoving state, we can observe from Table \ref{table4} that the Precision for the "unmoving" state reaches 94.092\%, the F1-score is 96.923\%, and the Recall is 99.974\%. In the confusion matrix shown in Figure 4, the values for false positives and false negatives are very low, indicating that the model performs exceptionally well in the binary classification task, particularly in applications that require high precision in determining "whether the vehicle is moving."
Overall, the binary classification approach is more suitable for practical applications. However, to accurately detect the vibration state, future work will focus on optimizing CLAHE enhancement parameters, adjusting the ROI region, or introducing more complex models to improve the performance of the three-class classification.
\begin{figure}[h]
\centering
\graphicspath{ {./} }
\includegraphics[width=2.5in]{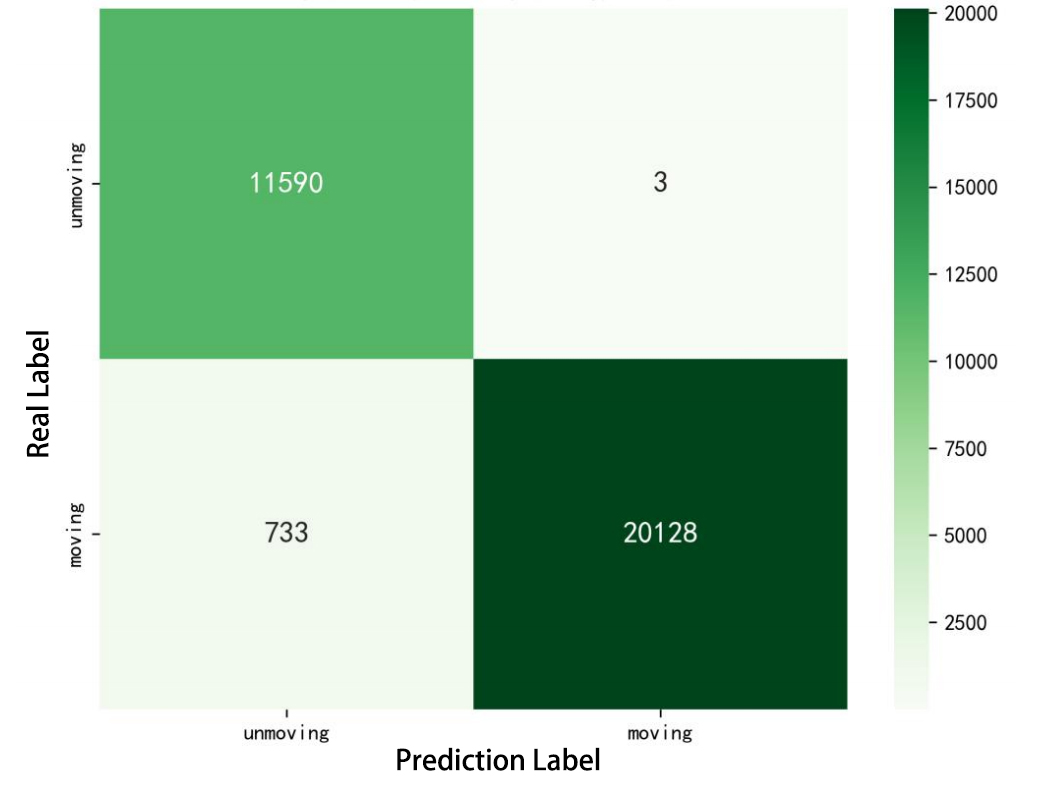}
\caption{Confusion matrix for classification of two types of motion states}
\label{fig5}
\end{figure}

\begin{table}[!h]
    \centering
    \caption{Performance analysis of two types of motion state classification details}
    \begin{tabular}{
        l               
        S[table-format=1.4]  
        S[table-format=1.4]  
        S[table-format=1.4]  
    }
    \toprule
    \textbf{State} & \textbf{Precision} & \textbf{Recall(Accuracy)} & \textbf{F1-score} \\
    \midrule
    
    unmoving & 0.94052 & 0.99974 & 0.96923 \\
    moving  & 0.99985 & 0.96486 & 0.98205 \\
    \bottomrule
    \end{tabular}
    \label{table4}
\end{table}

\textbf{Performance Testing}:
Performance testing evaluates the algorithm's processing efficiency by measuring latency and CPU utilization at different resolutions. The results are shown in Table \ref{table5}.

From Table \ref{table5}, it can be observed that at a resolution of 704×576 px, the algorithm's latency is 14.2±1.3 milliseconds, with a CPU utilization of 9.8\%. At a resolution of 1280×720 px, the latency increases to 22.7±2.1 milliseconds, and the CPU utilization is 14.3\%. This indicates that the algorithm can maintain low latency and low CPU utilization when processing images at different resolutions, demonstrating high processing efficiency.

In real-time vision processing tasks, low latency and high processing efficiency are key factors for ensuring system responsiveness. The algorithm proposed in this study performs well in this regard, meeting the real-time requirements of commercial vehicle AEB systems.

\begin{table}[ht]
    \centering
    \caption{The relationship between resolution and system performance}
    \begin{tabular}{
        c               
        S[table-format=2.1(2)]  
        S[table-format=2.1]     
    }
    \toprule
    \textbf{Resolution (px)} & \textbf{latency (ms)} & \textbf{CPU Usage （\%）} \\
    \midrule
    704×576   & 14.2 \pm 1.3 & 9.8 \\
    1280×720  & 22.7 \pm 2.1 & 14.3 \\
    \bottomrule
    \end{tabular}
    \label{table5}
\end{table}

\subsection{Field Implementation}

To validate the effectiveness of the vision-based trajectory analysis approach for zero-speed state detection in commercial vehicle AEB systems in real-world applications, we conducted a 6-month fleet test with 142 trucks. Before the test, the fleet frequently experienced braking errors due to inaccurate zero-speed detection, which impacted both transportation efficiency and safety. After deploying the system, the number of false braking events decreased by approximately 89\%. This improvement is attributed to the algorithm's precise identification of the vehicle’s motion state, preventing the AEB system from being falsely triggered when the vehicle is actually stationary or moving at low speeds.

During the testing period, when a vehicle suddenly braked or a pedestrian unexpectedly entered the roadway, the AEB system was able to trigger braking accurately and promptly. This demonstrated the reliability and effectiveness of the solution in real-world emergencies, ensuring the safety of both vehicles and pedestrians at critical moments.

Although the solution performed well in the overall testing, some failure cases still occurred. We conducted a detailed analysis of the causes of these failures and proposed corresponding mitigation strategies, as outlined in Table \ref{table6}.

\begin{table}[ht]  
    \centering  
    \caption{Fault Frequency Mitigation Strategies}  
    \begin{tabular}{lll}  
    \toprule  
    \textbf{Mitigation Strategy} & \textbf{Fault Cause} & \textbf{Frequency} \\  
    \midrule  
    Sensor and wheel encoder fusion & Extremely low texture & 2.1\% \\  
    Activated rain removal algorithm & Heavy rainfall & 1.7\% \\  
    Adaptive exposure control & Sudden lighting change & 1.2\% \\  
    \bottomrule  
    \end{tabular}  
    \label{table6}  
\end{table}

\section{Discussion}
\subsection{Contribution}
The main contributions of this study are as follows:

\textbf{High-Precision Zero-Speed State Detection}: This study introduces the SIFT-trajectory statistical method in the commercial vehicle zero-speed AEB scenario. By conducting a detailed analysis of the vehicle's motion trajectory, the method accurately distinguishes whether the vehicle is in a stationary or moving state, achieving a high detection accuracy of 99.97\%. This effectively prevents false braking caused by misjudgment.

\textbf{Effective Solution to the "0 km/h False Braking" Problem}: In a practical test, the occurrence of false braking events was reduced by 89\%. This demonstrates that the algorithm can accurately identify the vehicle's actual motion state, preventing the AEB system from being falsely triggered when the vehicle is stationary or moving at low speeds. This addresses the "0 km/h false braking" issue. In real-world applications, many commercial vehicles may experience rear-end collisions and other secondary accidents due to improper AEB system activation, which not only affects transportation efficiency but also endangers the driver’s safety. The application of this algorithm can effectively reduce these risks and improve road transportation safety.

\subsection{Limitations}
During the practical testing, we identified several issues:
\begin{enumerate}
\item\textbf{Insufficient Algorithm Robustness in Low-Texture and Extreme Weather Scenarios}: The algorithm's performance in environments with low texture or extreme weather conditions was not optimal, indicating the need for further improvements in these areas.

\item\textbf{Confusion Between Static and Vibrating States}: There is still some confusion between the static and vibrating states, as the distinction between them is relatively low. However, for practical applications, the focus is primarily on differentiating between moving and non-moving states. Future work will delve deeper into addressing this issue.

\item\textbf{Lack of Data for Extreme Weather Conditions}: The dataset used for testing lacks sufficient data for extreme weather scenarios. Future work will explore incorporating data from the Ithaca365 dataset [27] to include more extreme weather conditions for further training and refinement.
\end{enumerate}
  
\subsection{Future Work}
\textbf{Integrating Millimeter-Wave Radar to Enhance Performance in Rain and Fog}: Although the method proposed in this study performs well in most environments, the performance of vision sensors may be impacted to some extent in adverse weather conditions such as heavy rain and fog (\cite{ref29}). Radar technology has already been integrated into mass-produced vehicles for applications such as adaptive cruise control (ACC) and side-warning assistance (\cite{ref30,ref31,ref32,ref33}). Therefore, considering the use of radar is essential. LiDAR is widely used in autonomous vehicles for obstacle detection (\cite{ref34}); however, \cite{ref35} suggests that normal rain and heavy rain may affect LiDAR performance. \cite{ref36} highlights that millimeter-wave radar has the advantage of being unaffected by weather conditions and accurately detecting the distance, speed, and angle of objects in harsh weather such as rain and fog. Future work will integrate millimeter-wave radar technology. By fusing data from visual sensors and millimeter-wave radar, the strengths of both can be fully utilized (\cite{ref15,ref16}), improving the accuracy and reliability of vehicle state detection.

\textbf{Developing an FPGA Accelerated Version}: To meet the requirements of large-scale production and demonstrate the algorithm's processing efficiency and real-time performance, future work will also focus on developing a Field-Programmable Gate Array (FPGA) accelerated version. \cite{ref37,ref38} shows that FPGA offers greater flexibility, shorter time-to-market, reliability, and maintainability. Additionally, an FPGA's powerful parallel processing capability can accelerate the hardware implementation of the algorithm, improving operational speed. By porting the algorithm to an FPGA platform, real-time processing can be achieved, ensuring that the AEB system can make correct decisions instantly.

In summary, the vision-based trajectory analysis algorithm for zero-speed state detection in commercial vehicle AEB systems proposed in this study offers significant advantages and broad application prospects. Through continuous optimization and improvement, this algorithm is expected to play an important role in the field of commercial vehicle safety, contributing to the safety and efficiency of road transportation.

\section{Conclusion}
The vision-based trajectory analysis algorithm for zero-speed state detection in commercial vehicle AEB systems proposed in this study aims to reduce the issue of "zero-speed braking" false triggers, improving the safety, comfort, and passenger protection in commercial vehicle operations. The algorithm integrates SIFT feature extraction with trajectory statistics, and through theoretical innovation and experimental validation, it has demonstrated a significant breakthrough in addressing the "zero-speed braking" false trigger problem. This algorithm shows substantial advantages and broad application prospects. With continuous optimization and improvement, it is expected to play a crucial role in the field of commercial vehicle safety, contributing to the safety and efficiency of road transportation.

\section*{Conflicts of Interest}
The authors declare no conflicts of interest.

\section*{Funding}
The authors received no financial support for the research, authorship, and/or publication of this article.

\section*{Data Availability Statement}
The data that support the findings of this study are available from the corresponding author upon reasonable request.

\bibliographystyle{splncs04}
\bibliography{ref}

@techreport{ref1,
  author       = {A. Dhani},
  title        = {{Reported road casualties in Great Britain: Quarterly provisional estimates year ending September 2017}},
  institution  = {U.K. Department for Transport},
  address      = {London, UK},
  type         = {Technical Report},
  month        = feb,
  year         = {2018},
  url          = {https://assets.publishing.service.gov.uk/government/uploads/system/uploads/attachment_data/file/681593/quarterly-estimates-july-toseptember-2017.pdf}
 
}

@techreport{ref2,
  author       = {S. Singh},
  title        = {{Traffic safety facts}},
  institution  = {Nat. Highway Traffic Saf. Admin., U.S. Dept. Transp },
  address      = {Washington, DC, USA},
  number       = {DOT HS 812 115},
  month        = feb,
  year         = {2015},
  type         = {Tech. Rep.},
  url          = {https://crashstats.nhtsa.dot.gov/Api/Public/ViewPublication/812115}
}

@article{ref3,
  title={Research on the impacts of connected and autonomous vehicles (CAVs) on traffic flow},
  author={Atkins, WS},
  journal={Stage 2: Traffic modelling and analysis technical report},
  year={2016},
  publisher={Department for Transport London, UK}
}

@article{ref4,
  author    = {H. Zhang and D. Kong and G. Ao and others},
  title     = {Research progress of vehicle active collision prevention control technology},
  journal   = {High Tech. Lett.},
  volume    = {32},
  number    = {3},
  pages     = {314--326},
  year      = {2022}
}

@misc{ref5,
  title={False event suppression for collision avoidance systems},
  author={Trombley, Roger Arnold and Pilutti, Thomas Edward},
  year={2013},
  month=feb # "~5",
  publisher={Google Patents},
  note={US Patent 8,370,056}
}

@techreport{ref6,
  author      = {{Robert Bosch GmbH}},
  title       = {{CAN Specification Version 2.0}},
  institution = {Robert Bosch GmbH},
  address     = {Stuttgart, Germany},
  year        = {1991},
  type        = {Technical Report}
}

@article{ref7,
  title={Cart-mounted geolocation system for unexploded ordnance with adaptive ZUPT assistance},
  author={Li, Leilei and Pan, Yingjun and Lee, Jong-Ki and Ren, Chunhua and Liu, Yu and Grejner-Brzezinska, Dorota A and Toth, Charles K},
  journal={IEEE Transactions on Instrumentation and Measurement},
  volume={61},
  number={4},
  pages={974--979},
  year={2012},
  publisher={IEEE}
}

@article{ref8,
  title={A novel vehicle stationary detection utilizing map matching and IMU sensors},
  author={Amin, Md Syedul and Reaz, Mamun Bin Ibne and Nasir, Salwa Sheikh and Bhuiyan, Mohammad Arif Sobhan and Ali, Mohd Alauddin Mohd},
  journal={The Scientific World Journal},
  volume={2014},
  number={1},
  pages={597180},
  year={2014},
  publisher={Wiley Online Library}
}

@article{ref9,
  title={Zero-velocity detection—An algorithm evaluation},
  author={Skog, Isaac and Handel, Peter and Nilsson, John-Olof and Rantakokko, Jouni},
  journal={IEEE transactions on biomedical engineering},
  volume={57},
  number={11},
  pages={2657--2666},
  year={2010},
  publisher={IEEE}
}

@inproceedings{ref10,
  title={An improved SIFT feature extraction method for tyre tread patterns retrieval},
  author={Wang, Shuai and Liu, Ying and Li, Daxiang and Yan, Haoyang and Bai, Bendu},
  booktitle={2014 Seventh international symposium on computational intelligence and design},
  volume={1},
  pages={539--543},
  year={2014},
  organization={IEEE}
}

@article{ref11,
  title={Distinctive image features from scale-invariant keypoints},
  author={Lowe, David G},
  journal={International journal of computer vision},
  volume={60},
  number={2},
  pages={91--110},
  year={2004},
  publisher={Springer}
}

@inproceedings{ref12,
  title={Object recognition from local scale-invariant features},
  author={Lowe, David G},
  booktitle={Proceedings of the seventh IEEE international conference on computer vision},
  volume={2},
  pages={1150--1157},
  year={1999},
  organization={Ieee}
}

@inproceedings{ref13,
  title={S-SIFT: A Simple SIFT Algorithm with High Efficiency},
  author={Wang, Yixue and Huang, Yujie and Peng, Liyuan and Wang, Mingyu and Li, Wenhong and Jing, Minge and Zeng, Xiaoyang},
  booktitle={2024 IEEE 17th International Conference on Solid-State \& Integrated Circuit Technology (ICSICT)},
  pages={1--3},
  year={2024},
  organization={IEEE}
}

@article{ref14,
  title={Semantic foggy scene understanding with synthetic data},
  author={Sakaridis, Christos and Dai, Dengxin and Van Gool, Luc},
  journal={International Journal of Computer Vision},
  volume={126},
  number={9},
  pages={973--992},
  year={2018},
  publisher={Springer}
}

@inproceedings{ref15,
  title={Frontal object perception using radar and mono-vision},
  author={Chavez-Garcia, R Omar and Burlet, Julien and Vu, Trung-Dung and Aycard, Olivier},
  booktitle={2012 IEEE Intelligent Vehicles Symposium},
  pages={159--164},
  year={2012},
  organization={IEEE}
}

@inproceedings{ref16,
  title={Vision-based lateral position improvement of radar detections},
  author={Nishigaki, Morimichi and Rebhan, Sven and Einecke, Nils},
  booktitle={2012 15th International IEEE Conference on Intelligent Transportation Systems},
  pages={90--97},
  year={2012},
  organization={IEEE}
}

@manual{ref17,
  author      = {{NVIDIA}},
  title       = {{\textit{NVIDIA Jetson AGX Xavier}}},
  howpublished = {[Online]},
  url         = {https://www.nvidia.com/eses/autonomous-machines/embedded-systems/jetson-agx-xavier/},

}

@inproceedings{ref18,
  title={CosineRecom: A KNN-Based Movie Recommendation System Using Cosine Similarity},
  author={Singh, Mandeep and Mishra, Prafull and Aggarwal, Nitin and Kumar, Vinay and Sharma, Pramod Kumar and Yadav, Ritesh},
  booktitle={2024 4th International Conference on Advancement in Electronics \& Communication Engineering (AECE)},
  pages={190--194},
  year={2024},
  organization={IEEE}
}

@inproceedings{ref19,
  title={Adaptive color independent components based sift descriptors for image classification},
  author={Ai, Danni and Han, Xianhua and Ruan, Xiang and Chen, Yen-Wei},
  booktitle={2010 20th International Conference on Pattern Recognition},
  pages={2436--2439},
  year={2010},
  organization={IEEE}
}

@incollection{ref20,
  title={Scale-invariant feature transform (SIFT)},
  author={Burger, Wilhelm and Burge, Mark J},
  booktitle={Digital Image Processing: An Algorithmic Introduction},
  pages={709--763},
  year={2022},
  publisher={Springer}
}

@article{ref21,
  title={A survey on 3d object detection methods for autonomous driving applications},
  author={Arnold, Eduardo and Al-Jarrah, Omar Y and Dianati, Mehrdad and Fallah, Saber and Oxtoby, David and Mouzakitis, Alex},
  journal={IEEE Transactions on Intelligent Transportation Systems},
  volume={20},
  number={10},
  pages={3782--3795},
  year={2019},
  publisher={IEEE}
}

@article{ref22,
  title={Robustness-aware 3d object detection in autonomous driving: A review and outlook},
  author={Song, Ziying and Liu, Lin and Jia, Feiyang and Luo, Yadan and Jia, Caiyan and Zhang, Guoxin and Yang, Lei and Wang, Li},
  journal={IEEE Transactions on Intelligent Transportation Systems},
  year={2024},
  publisher={IEEE}
}

@article{ref23,
  title={Dehazenet: An end-to-end system for single image haze removal},
  author={Cai, Bolun and Xu, Xiangmin and Jia, Kui and Qing, Chunmei and Tao, Dacheng},
  journal={IEEE transactions on image processing},
  volume={25},
  number={11},
  pages={5187--5198},
  year={2016},
  publisher={IEEE}
}

@inproceedings{ref24,
  title={GROUNDED: The Localizing Ground Penetrating Radar Evaluation Dataset.},
  author={Ort, Teddy and Gilitschenski, Igor and Rus, Daniela},
  booktitle={Robotics: Science and Systems},
  volume={2},
  year={2021}
}

@article{ref25,
  title={Canadian adverse driving conditions dataset},
  author={Pitropov, Matthew and Garcia, Danson Evan and Rebello, Jason and Smart, Michael and Wang, Carlos and Czarnecki, Krzysztof and Waslander, Steven},
  journal={The International Journal of Robotics Research},
  volume={40},
  number={4-5},
  pages={681--690},
  year={2021},
  publisher={SAGE Publications Sage UK: London, England}
}

@article{ref26,
  title={The apolloscape open dataset for autonomous driving and its application},
  author={Wang, Peng and Huang, Xinyu and Cheng, Xinjing and Zhou, Dingfu and Geng, Qichuan and Yang, Ruigang},
  journal={IEEE transactions on pattern analysis and machine intelligence},
  volume={1},
  year={2019}
}

@inproceedings{ref27,
  title={Ithaca365: Dataset and driving perception under repeated and challenging weather conditions},
  author={Diaz-Ruiz, Carlos A and Xia, Youya and You, Yurong and Nino, Jose and Chen, Junan and Monica, Josephine and Chen, Xiangyu and Luo, Katie and Wang, Yan and Emond, Marc and others},
  booktitle={Proceedings of the IEEE/CVF Conference on Computer Vision and Pattern Recognition},
  pages={21383--21392},
  year={2022}
}

@inproceedings{ref29,
  title={Research on Verification and Evaluation Method of Automatic Emergency Braking Function in Rain and Fog Weather Based on TOPSIS},
  author={Zhang, Jiangmin and Zhang, Qiang and Hu, Mengxia and Tang, Yu and Ma, Guosheng},
  booktitle={2024 10th International Conference on Control, Automation and Robotics (ICCAR)},
  pages={21--27},
  year={2024},
  organization={IEEE}
}

@article{ref30,
  title={Amplitude-modulated laser radar for range and speed measurement in car applications},
  author={Mao, Xuesong and Inoue, Daisuke and Kato, Satoru and Kagami, Manabu},
  journal={IEEE Transactions on Intelligent Transportation Systems},
  volume={13},
  number={1},
  pages={408--413},
  year={2011},
  publisher={IEEE}
}

@inproceedings{ref31,
  title={Electronically scanned millimeter-wave radar for pre-crash safety and adaptive cruise control system},
  author={Tokoro, S and Kuroda, K and Kawakubo, A and Fujita, K and Fujinami, H},
  booktitle={IEEE IV2003 Intelligent Vehicles Symposium. Proceedings (Cat. No. 03TH8683)},
  pages={304--309},
  year={2003},
  organization={IEEE}
}

@article{ref32,
  title={Looking at vehicles on the road: A survey of vision-based vehicle detection, tracking, and behavior analysis},
  author={Sivaraman, Sayanan and Trivedi, Mohan Manubhai},
  journal={IEEE transactions on intelligent transportation systems},
  volume={14},
  number={4},
  pages={1773--1795},
  year={2013},
  publisher={IEEE}
}

@inproceedings{ref33,
  title={Towards fully autonomous driving: Systems and algorithms},
  author={Levinson, Jesse and Askeland, Jake and Becker, Jan and Dolson, Jennifer and Held, David and Kammel, Soeren and Kolter, J Zico and Langer, Dirk and Pink, Oliver and Pratt, Vaughan and others},
  booktitle={2011 IEEE intelligent vehicles symposium (IV)},
  pages={163--168},
  year={2011},
  organization={IEEE}
}

@inproceedings{ref34,
  title={Multilayer lidar-based pedestrian tracking in urban environments},
  author={Sato, Seiichi and Hashimoto, Masafumi and Takita, Manabu and Takagi, Kiyokazu and Ogawa, Takashi},
  booktitle={2010 IEEE Intelligent Vehicles Symposium},
  pages={849--854},
  year={2010},
  organization={IEEE}
}

@inproceedings{ref35,
  title={The influence of rain on small aperture LiDAR sensors},
  author={Fersch, Thomas and Buhmann, Alexander and Koelpin, Alexander and Weigel, Robert},
  booktitle={2016 German Microwave Conference (GeMiC)},
  pages={84--87},
  year={2016},
  organization={IEEE}
}

@inproceedings{ref36,
  title={Automotive mm-wave radar: Status and trends in system design and technology},
  author={Wenger, J},
  booktitle={IEE Colloquium on Automotive Radar and Navigation Techniques},
  pages={1--1},
  year={1998},
  organization={IET}
}

@article{ref37,
  title={Choosing the right architecture for real-time signal processing designs},
  author={Adams, Leon and Marketing, Strategic},
  journal={Texas Instruments, Document Number SPRA879},
  year={2002}
}

@inproceedings{ref38,
  title={An FPGA-based system for real-time electrocardiographic detection of STEMI},
  author={Karim, Mohammed and El Kouache, Mustapha and Amarouch, Mohamed-Yassine and others},
  booktitle={2016 2nd International Conference on Advanced Technologies for Signal and Image Processing (ATSIP)},
  pages={830--835},
  year={2016},
  organization={IEEE}
}

\end{CJK}
\end{document}